# VIDEO-BASED DETECTION OF SQUINT AND CATARACT FOR ACCESSIBILITY-AWARE ADAPTIVE WEB INTERFACE RENDERING

Amar Ranjan Dash [1] and Manas Ranjan Patra [2]

[1]Department of Computer Science, Berhampur University, Berhampur, India
[2]CSE Department, NIST University, Pallur Hills, Berhampur, India

***ABSTRACT***

*Squint and cataract are major ocular disorders that majorly affect visual perception and interaction capability. This paper proposes a real-time video-based automated detection system for squint and cataract detection based on computer vision and image processing methods. The proposed system uses a media-pipe face-mesh (a 478-point facial landmark detection model) to extract geometric ocular features for multi-class squint classification. Simultaneously, The presence and severity cataract is estimated through grayscale intensity and histogram-based lens opacity analysis. The system records short video sequences with standard laptop or mobile cameras, which can be deployed at low costs and on a large scale. The experimental performance has shown great accuracy in the detection of squint (98.39%) and classification of cataract (96.90%). Besides automatic ocular analysis, the proposed framework is also made accessible for visual impairment inference which will be integrated with future adaptive user-interface and Web accessibility systems for people with visual impairment.*

## 1. INTRODUCTION

The eye is a sense organ which processes light and perceives visual and depth information. Any structural or functional defect of the eye can cause visual deficiency like squint, cataract, blur and colour blindness. Squint occurs due to abnormal ocular alignment. It means one or both eyes deviate from their anatomically coordinated positions. Squint results in impaired binocular vision and reduced depth perception. Cataract develops when crystallin proteins begin to clump together in the lens causing the lens to become opaque, causing problems with vision, glare and contrast perception. Typical diagnosis of the disorders involves specialized ophthalmic equipment and expert clinical evaluation, which are not readily available in low-resource areas.

The application of automated techniques such as image and video analysis are now possible for the identification of ocular disease, due to recent advances in computer vision and medical image processing. However, most existing approaches focus on isolated disease detection using static images and provide limited support for real-time multi-condition analysis.

This paper presents a hybrid approach for the simultaneous detection of squint and cataract based on geometric and intensity based analysis of the various features of the eye from live video. Media-pipe face-mesh (a 478-point facial landmark model) is used to locate the pupils and classify the image into multiple classes of squints using the facial landmark model, and the severity of cataract in the treated image is estimated by grayscale histogram and lens opacity analysis. The proposed system operates in real time using standard cameras. Additionally it supports accessibility-aware visual impairment inference for adaptive human-computer interaction systems.

## 2. LITERATURE REVIEW

Squint occurs when two pupils are misaligned and look in different direction, preventing coordinated binocular vision. Squint occurs when one pupil deviates inward, outward, upward or downward relative to the other resulting in misalignment and impaired binocular vision. First A R Elkington and P T Khaw [01] have analyzed different type of squint and mentioned different type of clinical test need to be conducted for detection of six type of squint. S. Yadav [02] have developed an algorithm for analysing age related macular degeneration in retina. G. Lennerstrand [03] have analyze the structure and function of eye muscles. He have analyzed the MRI image of eye muscles to detect disfunction of which ocular muscle leads to which type of strabismus. R. M. Potdar et al. [04] analyzed different type of squint and proposed a logical approach for the detection of squint related disorders. In [05], they further extended their previous work by implementing a MATLAB code, that utilizes hough transform to detect ocular misalignment indicative of squint from static facial images. S. Yadav [06] have proposed an image-based squint detection approach utilizing the hough transform in combination with a projection function to identify ocular misalignment. S. Tak, and K. J. Satao [07, 08] have developed a Graphical User Interface using python programming and "Support Vetor Machine" to detect six different types of squint from facial image. They further progress their research work in [09] by integrating image processing tool to extract some eye related parameters for enhanced squint analysis. M. Yarkheir et. al. [10] have developed a deep learning model for squint detection from facial image. D. Wu et. al. [11] have programmed a Visual Transformer "VIT_16_224" model with deep learning to detect the squint from near face image. W. Kurdthongmee et. al. [12] have developed a machine learning model for strabismus detection from monocular eye images in telemedicine application. M. R. Akbari [13] have performed a survey on Thyroid eye disease related strabismus. S Joshi [14] have develop a python programming based Graphical User Interface. This GUI use Convolutional Neural Network model to detect the type of squint based on uploaded image. B. J. Rea [15] have proposed an automated squint detection system for rats. J. Lu et. al. [16] have developed a strabismus detection system using Residual-Fully-Convolutional-Neural-Network tailored for telemedicine applications. N. S. Zolkifli, and A. Nazari [17] have developed a strabismus detection method based on Sobel-Edge-Detection and Hough-Transform. S. Karaaslan et. al. [18] have developed a squint detection system combining SVM classification with Hirschberg test. A. Sharna et, al. [19] have performed a survey on different existing squint detection techniques. G. F. Papalia et. al. [20] also performed a thorough survey on multiple strabismus detection system. They analysed the effect of strabismus on postural control of human being. They also try to analyze the effectiveness of realignment surgery, by conducting the strabismus detection before and after the surgery. Z. Zhao et. al. [21] developed a wearable googles for eye tracking and AI based Ocular feature analysis. This googles can be used for strabismus detection. Z. Yang et. al. [22] have performed a systematic review on methodology of twenty five plus strabismus detection techniques.

If any pupil becomes cloudy due to accumulation of protein layer, then it may lead to myopia at starting and blindness at later stage. M. Yang et. al. [23] pro-posed a neural-network-based classifier to classify the cataract, based on the retinal image. The classifier consists of preprocessing, feature extraction and classification. L. Zhang et. al. [24] and M. S. Junayed et. al. [25] have proposed techniques for cataract detection from fundus images using Deep

Convolutional Neural Network. M. Caixinha et. al. [26] have developed a model to detect the cataract of animal from their ultrasound based fundus image. I. Khan et. al. [27] have developed a ResNet-50 model for cataract detection from fundus-retinal images. J. J. Yang et. al. [28] developed an ensemble learning based model to classify the fondus retinal image and detect the presence of cataract. Similarly, H. Wu et. al. [29] programmed a multi-task learning model for cataract detection from the retinal fundus image. C. Muchibwa et. al. [30] conducted a survey on multiple fundus-eye image-based cataract detection technique. S. M. Khan et. al. [31] have developed a VGG-19 model for detecting the cataract from the fundus eye image. D. Patil et. al. [32] have performed a survey on nine different cataract detection techniques.

Jindal et. al. [33] developed a system to detect the cataract from eye images using digital image processing techniques and two distinct algorithms. The first algorithm employs histogram processing for feature extraction, analyzing intensity distributions to identify cataract induced changes in the lens. The second algorithm utilizes an area calculation method to quantify the extent of the cataractous region within the pupil, aiding in severity assessment. S. Agustin et. al. [34] have proposed a system to detect cataract from eye image using Convolutional Neural Network. H. M. Lopez et. al. [35] have performed a survey on different cataract detection techniques. They also classify different cataract detection techniques into different groups based on multiple parameters. P. Tripathi et. al. [36] have proposed a system for cataract detection from near infrared eye images using iris pupil segmentation and multi-task learning. T. Ganokratanaa [37] have programmed a “LeNet-Convolutional Neural Network” for detection of cataract from eye images. C. J. Lai et. al. [38] develop CNNDCI to detect the existence of cataract from close monocular eye image. J. Rana et. al. [39] try to develop an android application for cataract detection from close monocular eye image.

Kartynnik et. al. [40] introduced a real-time monocular dense 3D facial landmark reconstruction framework that was later developed into Media-Pipe Face Mesh. M. R. Patra et. al. [41,42] have evaluated the accessibility standard of 15 Indian web-portals and 15 web-portals of ASIA. Patra and Dash [43, 44] have analysed the accessibility standard of user-agents and its effect on accessibility of web-components. Motivated by accessibility studies reported by Patra and Dash [43, 44], the present work proposes a real-time web-came based framework that uses media-pipe facial land-marks and geometric position calculation for accessibility oriented detection of squint and cataracteye.

The rest of the paper organized as follows: Section 3.1 and 3.2, briefly elaborate on the concept of Squint and cataract respectively. Section 4, represents the squint and cataract detection methodology. Section 5, represents the result analysis of the proposed system and its comparison with traditional and existing approaches. Section 6, represents the concluding remarks and outlines the scope for future improvement.

## 3. BACKGROUND ON ACCESSIBILITY-RELEVANT OCULAR CONDITIONS

### 3.1. Squint (Strabismus)

The eye is a light-sensitive sensory organ responsible for vision. In order to perceive the image correctly, both pupil need to be aligned to send their images to the brain together so that the brain can produce a three dimensional (3D) image. Misalignment of the pupils can lead to squint (also known as strabismus). In other words squint occurs when one pupil deviates inward, outward, upward, or downward relative to the other one. Squints results in impaired binocular vision and reduced depth perception.

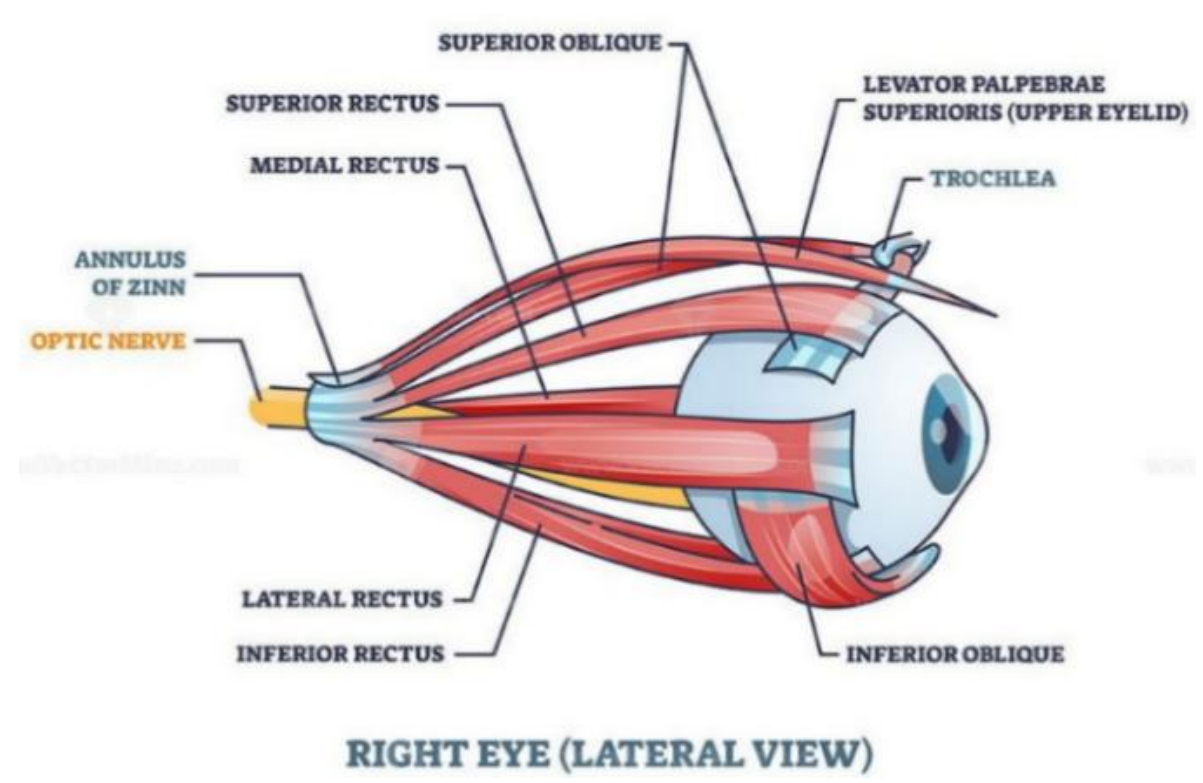

Figure 1: Extra-ocular muscles of human eye

Brains neurological signals controls the coordinated movement of both pupils through six extraocular muscles as represented in Figure 1. Damage to these muscles, associated nerves, or neuromuscular coordination mechanisms can disrupt synchronized eye movement and lead to squint. Damage to these muscles, associated nerves or mechanisms of neuromuscular coordination can be responsible for the absence of coordinated eye movements and a squint may also result. Squint can be classified into twelve types (as represented in Figure 2) depending on the direction and the type of deviation.

Squint impacts not just the way the eye looks, but also its function, leading to visual fatigue, unstable focus and trouble performing reading or digital interaction jobs. Most of the Traditional squint detection methods mainly depend on digital image analysis techniques like pupil localization, corneal reflection analysis, gaze estimation, and facial landmark detection. But. Most existing systems are limited to identifying only a few major squint categories. They primarily focus on disease detection without addressing accessibility-related visual interaction challenges.

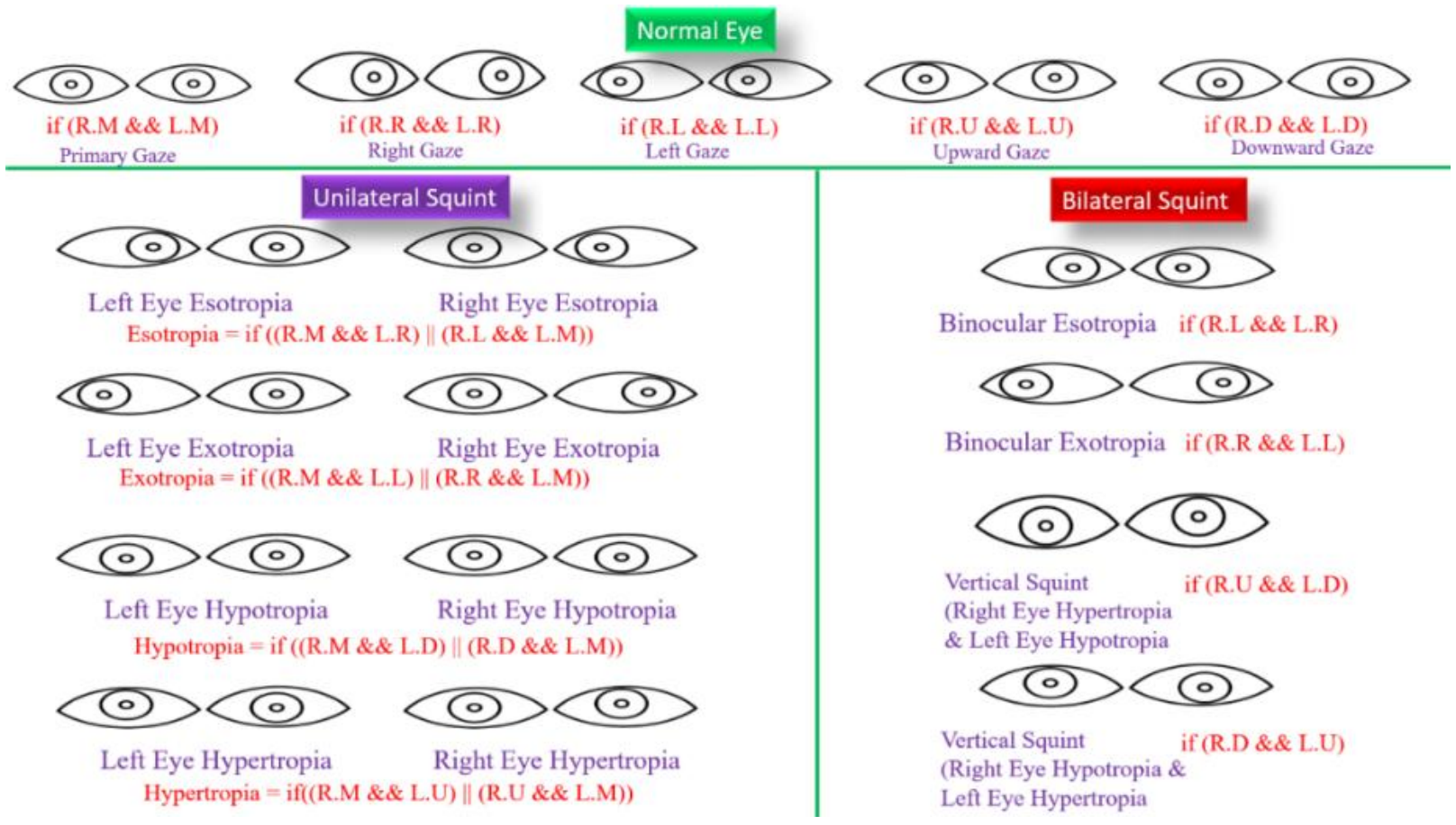

Figure 2: Type of Squint

## 3.2. Cataract

The pupil is a black opening located at the centre of the eye. It is is surrounded by a circular membrane known as the iris. The iris muscles control the diameter of the pupil by how much they dilate or contract the eye to allow more or less light to enter the eye. The crystalline lens is located behind the pupil, and is primarily composed of proteins known as crystallins. These proteins are necessary to keep the lens transparent and focus the light rays onto the retina so that they can be properly perceived.

Cataract is the process of crystallin proteins starting to clump together in certain parts of the lens, making it cloudy or opaque and blocking the passage of light. This opacity slowly impairs visual acuity, resulting in blurred vision, sensitivity to light, decreased contrast perception and loss of fine detail. Based on the location of protein deposition, cataracts are commonly classified into three major categories: nuclear cataract, cortical cataract, and posterior sub-capsular cataract. A cataract can have a major impact on digital visual interactions, especially when reading, particularly in high-luminance web environments. Cataract users may find it hard to read or see elements with a low contrast ratio; their vision might become unclear or blurry; and they may find it hard to use interface elements.

## 4. Proposed Methodology

The proposed framework simultaneously detects squint and cataract by means of facial land mark detection with media-pipe, geometric analysis of ocular landmarks and estimation of lens opacity based on intensity. The entire architecture can be broken down into four stages: image acquisition, pre-processing, feature extraction, and disease classification. The proposed system is able to detect multiple types of ocular alignment abnormalities and the severity of the cataract in both static images and real-time video streams. The overall process of the proposed methodology is shown in Figure 3.

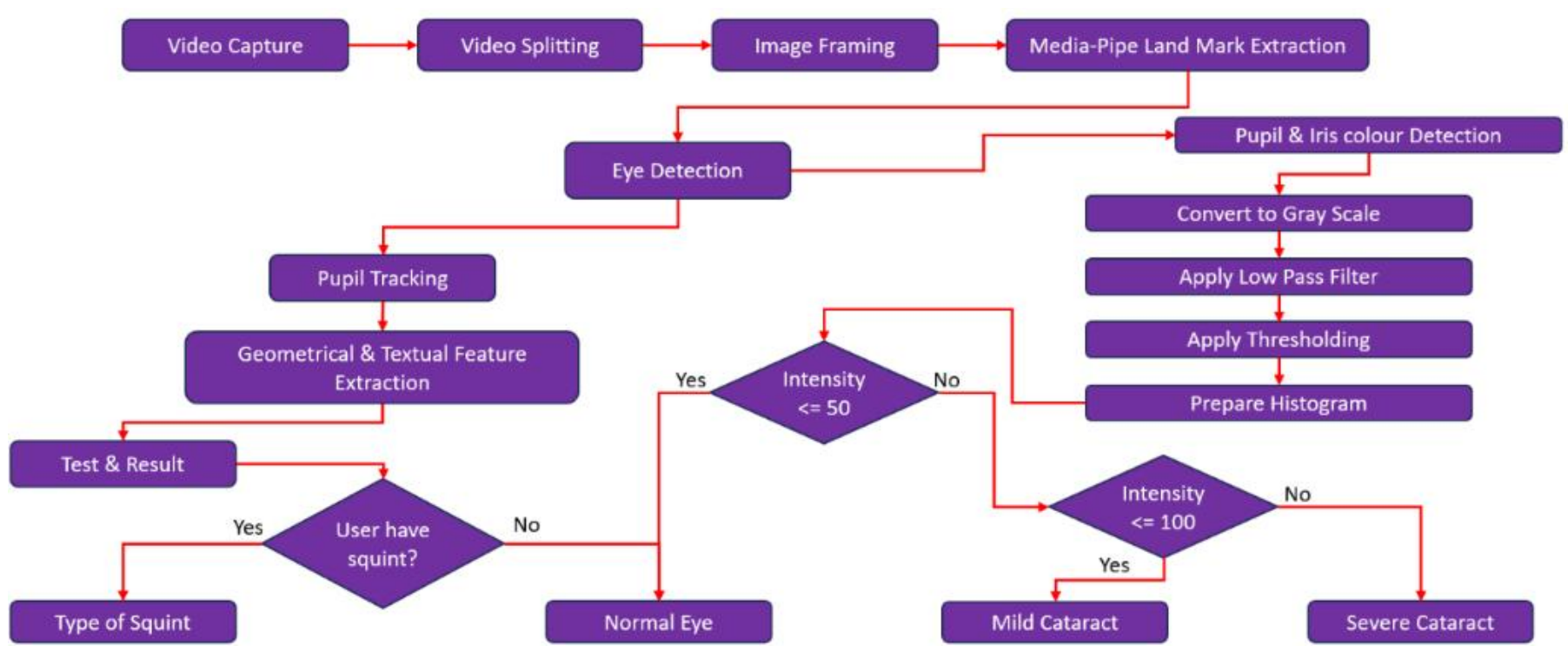

Figure 3: Methodology

### 4.1. Media-Pipe Face Mesh

*Media-Pipe Face Mesh is 478-point facial landmark detection model shown in* Figure *4*. It is used in the localization of the regions of the eye, which allows for the correct extraction of the pupil coordinates and the outline of eye shape. The displacement of the pupil with respect to the centres of the eyes is used for the classification of squints and the intensity variation of the optical intensity in the lens region is used for the detection of cataract.

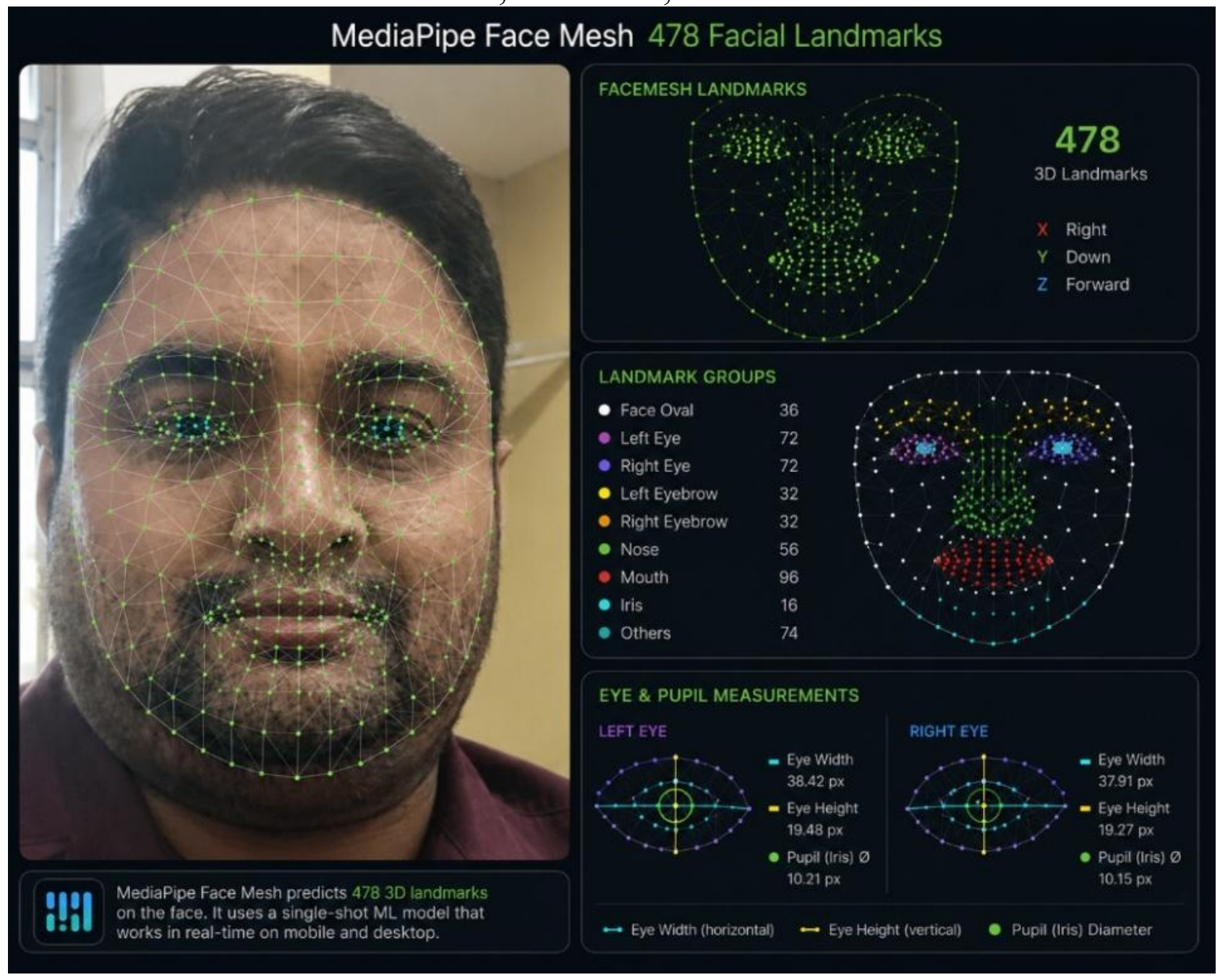

Figure 4: Media-Pipe Face-mesh Landmarks

## 4.2. Squint Classification Framework

The movement of each pupil is controlled by six extra-ocular muscles which enable precise movements as shown in
Figure *1*. Variations in muscle coordination or neural control can lead to deviations in pupil alignment, forming distinct ocular states.Considering the relative spatial orientation of both pupils, 17 different ocular configurations are defined in this study, which are displayed in
Figure *2*.Among these,five are normal configurations and the other twelve are various types of squint (strabismus).These conditions are derived from directional deviations of one or both eyes from their anatomically neutral positions.The squint can be classified into four unilateral and four bilateral types of squint as follows:

- Unilateral Squint (Single-eye deviation):
  - Esotropia: inward (nasal) deviation of the pupil
  - Exotropia: outward (temporal) deviation of the pupil
  - Hypertropia: upward deviation of the pupil
  - Hypotropia: downward deviation of the pupil
- Bilateral Squint (Both-eye involvement):
  - Binocular Esotropia: inward deviation of both pupils
  - Binocular Exotropia: outward deviation of both pupils
  - Vertical Misalignment: asymmetric vertical displacement between both pupils

These configurations are determined using geometric relationships between pupil locations and anatomical eye landmarks extracted from facial geometry.

## 4.3. Training Phase

During training, the model learns two complementary feature spaces:

- **Geometric Feature Space (Squint Analysis):** Derived from spatial coordinates of pupil positions relative to eye landmarks. These features capture horizontal and vertical deviations indicative of ocular misalignment.
- **Photometric Feature Space (Cataract Analysis):** Derived from intensity distributions within the pupil and lens region. These features reflect variations in opacity caused by protein aggregation in the crystalline lens.

Prior to feature extraction, input images are enhanced using a low-pass filtering operation to suppress high-frequency noise and improve landmark stability. The extracted feature vectors are then used to train a Support Vector Machine (SVM), selected for its robustness in handling non-linear decision boundaries in high-dimensional feature spaces.

## 4.4. Testing Phase

During inference, real-time video input is processed through a sequential pipeline:

a. **Frame Decomposition:** The input video stream is segmented into individual frames for independent analysis.
b. **Facial Landmark Localization**: Each frame is processed using a 478-point landmark detection model to extract eye contours and pupil positions.
c. **Squint Feature Computation and Rule-Based Classification:** Pupil coordinates are mapped to predefined spatial regions relative to the eye. Ten positional indicators are defined:

- R.R: Right pupil at lateral extreme
- R.M: Right pupil at centre
- R.L: Right pupil at medial side
- R.U: Right pupil in superior region
- R.D: Right pupil in inferior region
- L.R: Left pupil at medial side
- L.M: Left pupil at centre
- L.L: Left pupil at lateral extreme
- L.U: Left pupil in superior region
- L.D: Left pupil in inferior region

The combination of these indicators is evaluated using deterministic rule sets (as defined in Figure \ref{Fig2}) to classify ocular alignment into normal or squint categories.

d. **Cataract Detection Pipeline:** The eye region is converted from RGB to grayscale to emphasize intensity variations. A low-pass filter is applied to reduce noise, followed by histogram analysis of pixel intensity distribution within the lens region. The degree of lens opacity is estimated based on intensity thresholds:

- 0–50: Healthy
- 51–100: Mild Cataract
- 101–255: Severe Cataract

## 4.5. System Output

The final system output is obtained by combining results from both pipelines. Squint classification is determined from geometric deviation patterns, while cataract severity is inferred from intensity-based optical degradation. The overall detection process is summarized through Figure 3 and Squint-Cataract-Detection Algorithm.

**Squint-Cataract-Detection**

Input: Video streamRecV
Output{Squint classification and cataract level}

Step-1. Segment RecVs$ into individual frames;
Step-2. **for** each frame fdo
Step-3. Detect 478 facial landmarks;
Step-4. Extract geometric features from pupil positions;
Step-5. Classify squint type using deterministic rules;
Step-6. Convert pupil region into grayscale;
Step-7. Apply low-pass filter and extract histogram;
Step-8. Classify cataract level based on intensity:
Step-9. **if**{score ∈ [0--50]}
Step-10. Healthy eye;
**Step-11.** **else**
Step-12. **if**{score $\in$ [51--100]}
Step-13. Mild cataract;
**Step-14.** **else**
Step-15. Severe cataract;

## 4.6. Dataset Description

This study utilizes combination of three datasets comprising both images and videos for the detection of squint and cataract conditions. Image data set of overall 1,000 images are sourced from two publicly available dataset of Kaggle. Image dataset [45, 46] consist of multiple images representing different ocular conditions under diverse imaging conditions. Additionally, a proprietary data set comprising 1,000 video recordings was systematically acquired within the premises of the authors' affiliated institutions, namely "Parala Maharaja Engineering College" and "Berhampur University". The videos were recorded under controlled ambient lighting using standard laptop webcams during the period of January 2026 to March 2026. Each video captures a frontal view of the subject's face for approximately 10 seconds, which makes frame-by-frame analysis possible. The combined datasets include visual data of 2,000 persons. It contains iamge and video of 1,100 males and 900 females, ranging in age from 16 to 65 years. This heterogeneous demographic distribution enhances the model's robustness and generalizability across diverse patient populations.

## 4.7. Implementation

The proposed model identifies the type of squint, based on geometrical location of left and right pupil with respect to respective eye. Simultaneously, the proposed model will detect the presence of cataract, based on the intensity of pupil's pixel. Figure 5 illustrates the model's diagnostic outcomes across several test cases:

- Figure 5(a): the person in figure exhibits normal ocular alignment with no abnormalities.
- Figure 5(b): Diagnosis indicates Exotropia in the left eye, characterized by lateral outward deviation.
- Figure 5(c): The person presents Esotropia in the right eye, identified by medial inward deviation.
- Figure 5(d): Detection of Hypertropia in the left eye, evident from the superior vertical displacement of the pupil.
- Figure 5(e): Hypotropia diagnosed in the right eye due to inferior displacement.
- Figure 5(f): Bilateral Esotropia is detected, indicating convergent squint in both eyes.
- Figure 5(g): Severe cataract is diagnosed in the right eye, based on significant opacity inferred from reduced pupil pixel intensity.
- Figure 5(h): Asymmetric cataract severity is observed—mild in the left eye and severe in the right eye.

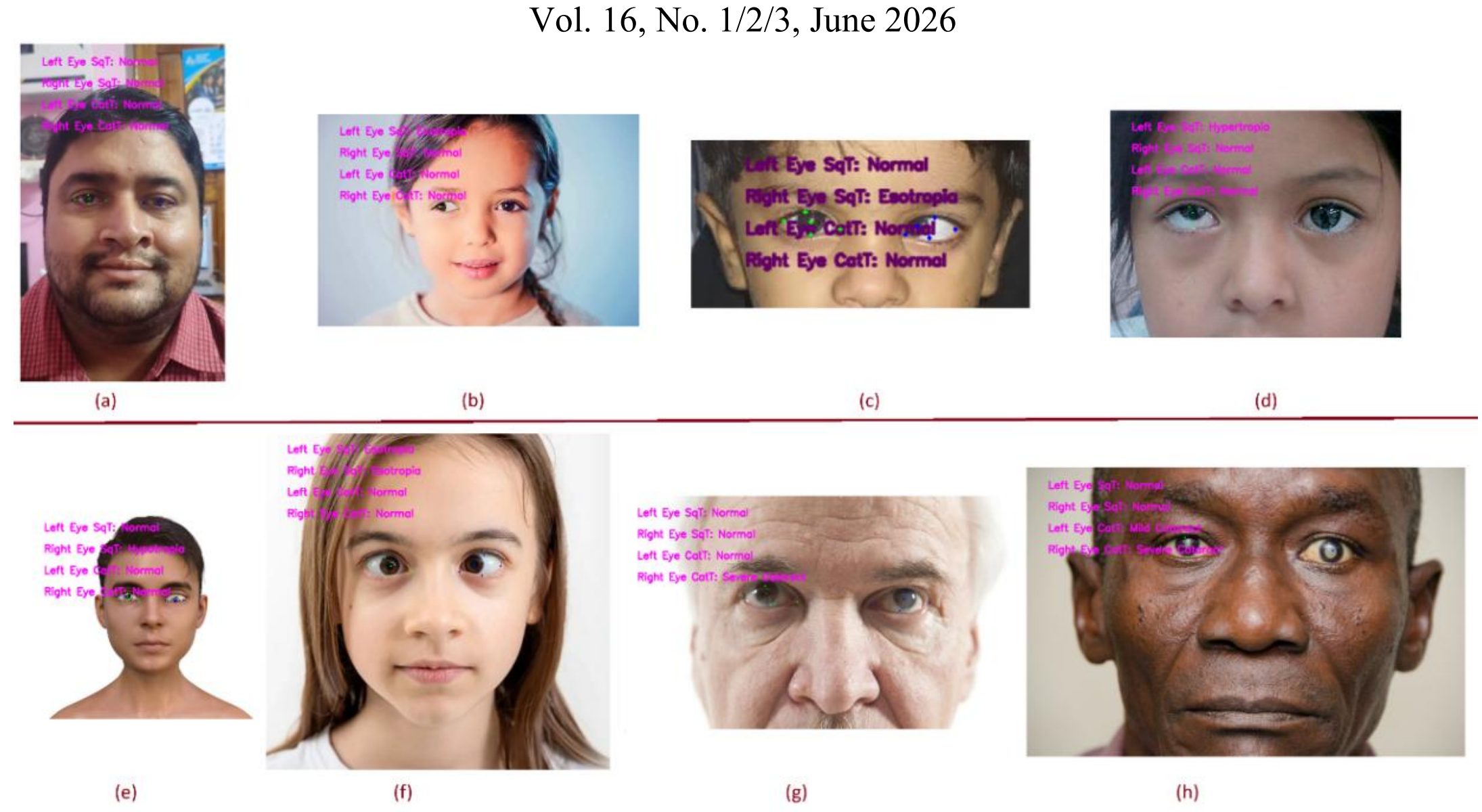

Figure 5: Model's results for some users

# 5. Result Analysis

The proposed framework illustrates real-time diagnosis of squint and cataract condition in the eye by means of live video based ocular analysis pipeline. The system combines intensity-based lens opacity analysis with geometric eye alignment to provide high classification accuracy across multiple ocular conditions. The proposed model is robust enough and has been tested experimentally across a range of imaging conditions, head orientations and illumination environments.

## 5.1. Squint Detection Performance

The proposed model outperformed the other models with an accuracy of 98.39%, precision of 99.24% and recall of 97.83% for the classification of squint. The results showed that the proposed geometric feature extraction technique was effective for detecting unilateral and bilateral squint. High precision value signifies a good model to reduce false-positive classifications, and a high recall confirms good detection of various deviation patterns of the eye.

A 478 point facial landmark detection framework is effectively used to achieve high accuracy in pupil localization and spatial alignment analysis. The proposed rule-based geometric classification mechanism is able to categorize a number of different categories of squint (esotropia, exotropia, hypertropia, hypotropia and binocular variants). The classification results and the confusion matrix are shown in Figure 6, which shows that the model has good inter-class separation and generalization. Table 1 shows a comparison of the proposed approach with the conventional methods of squint detection.

Besides clinical inference, correct squint detection helps in the accessibility-aware interface adaptation. Any deviation from alignment can lead to visual fatigue, loss of focus and decreased continuity of reading when using a digital device. Specific squint conditions can be identified and the proposed system can assist with adaptive accessibility mechanisms, such as optimized spacing, reduction of visual clutter, stabilization of reading layouts, and guidance of focus alignment for the visually impaired user.

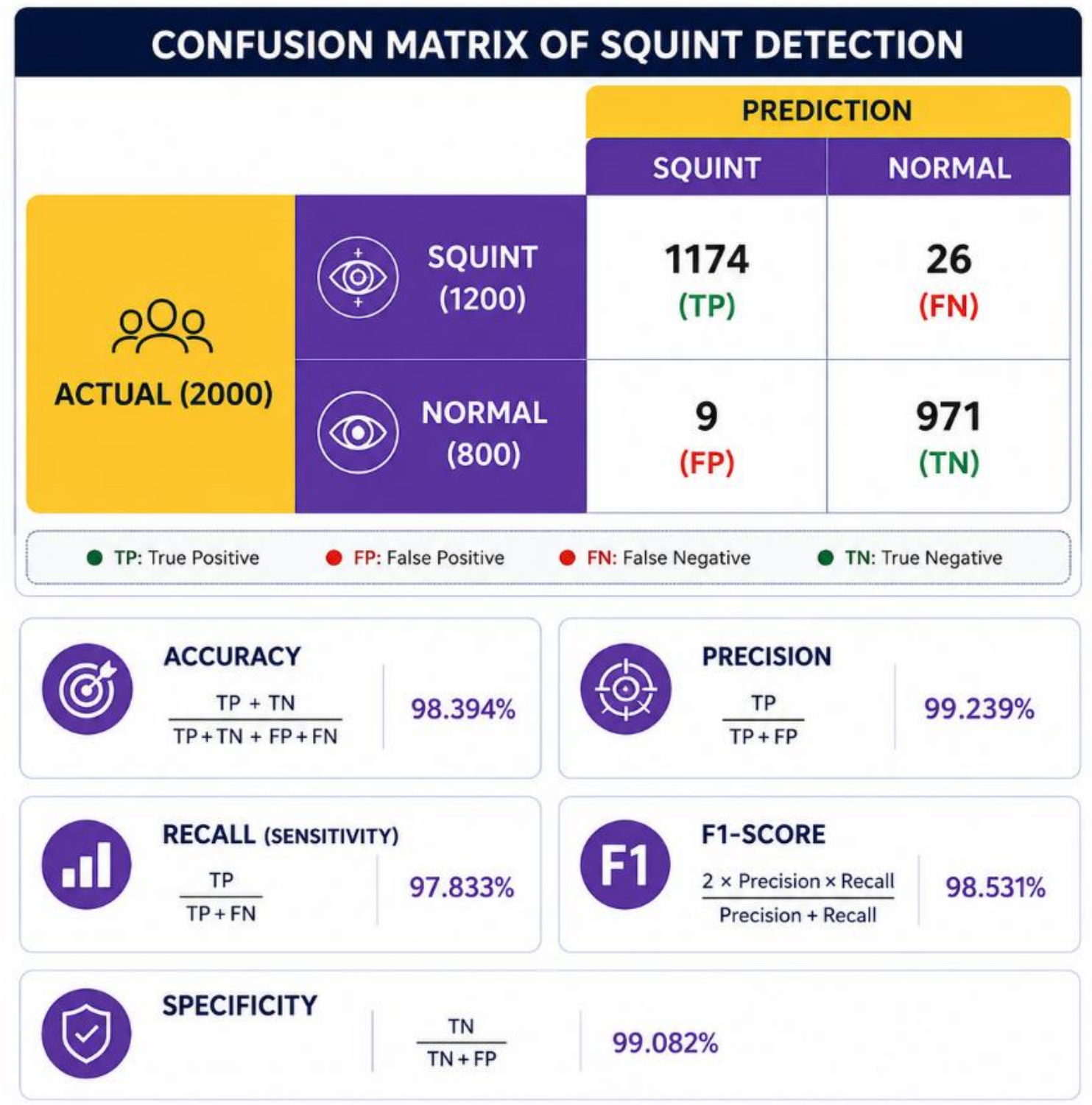

Figure 6: Performance of the model in Squint detection

Table 1: Quantitative Comparison of Squint detection Techniques

| Reference | Methodology Used | Application | Accuracy |
|---|---|---|---|
| S. Yadav [02] | Analysis of age-related macular degeneration in retina | | |
| R. M. Potdar et al. [04] | Proposed a logic using image Processing | | |
| R. M. Potdar et al. [05] | MATLAB code using Hough Transform and projection function | Squint detection from image | no confusion matrix |
| S. Yadav [06] | Proposed an approach using Hough Transform + Digital image processing + projection function | | |
| S. Tak, and K. J. Satao [07] | Support Vector Machine + Geometrical Location | Six type of Squint detection from image | Accuracy 85% |
| S. Tak, and K. J. Satao [08] | Support Vector Machine + Geometrical Location | Six type of Squint detection from image | Accuracy 85% |
| S. Tak, and K. J. Satao [09] | Support Vector Machine + Geometrical Location | Six type of Squint detection from image | Accuracy 85% |
| M. Yarkheir et. al. [10] | Deep Learning Model | Sour type of squint detection | Accuracy: 92% |
| D. Wu et. al. [11] | Visual Transformer (VIT_16_224) model + Deep Learning | Squint detection from image | Accuracy: 98% |
| W. Kurdthongmee et. al. [12] | Machine Learning Model | Squint detection from monocular eye images | No Confusion Matrix |

| M. R. Akbari [13] | Survey on thyroid eye disease related strabismus | | |
|---|---|---|---|
| S Joshi [14] | Python Programming based Graphical User Interface + Convolutional Neural Network | Six type of squint detection from facial image | No confusion Matrix |
| B. J. Rea [15] | drug administration (formalin, CGRP, formalin) + video mining | Video based squint detection in mouse | no confusion matrix |
| J. Lu et. al. [18] | Eye-region segmentation + geometrical location of pupil | six type of squint detection from image | Accuracy: 93.89% Specificity: 96.17% Sensitivity: 93.30% |
| N. S. Zolkifli, and A. Nazari [19] | Image mining + Hough Transform | Squint detection from image | no confusion matrix |
| S. Karaaslan et. al. [20] | Image processing + facial land mark + Hirschberg Test | Squint detection from high resolution images captured | Accuracy 90.50% |
| A. Sharna et, al. [21] | Survey on Ten different squint detection technique | | |
| Z. Yang et. al. [24] | Survey on Twenty five different squint detection technique | | |
| G. F. Papalia et. al. [22] | 1. Survey on squint detectiontechnique2. Effect of Strabismus onpostural control | | |
| Z. Zhao et. al. [23] | Wearable Googles | Strabismus Detection | Accuracy: 97% |
| Proposed Model | Python code based video mining using:<br>**1.** 478 based facial land mark detection<br>**2.** Geometrical positioning of pupil w.r.t land mark of respective eye<br>**3.** squint deterministic rules mentioned in figure 2. | **1.** all 12 type of squint and fivetype of normal condition canbe detected from live videos<br>**2.** cheapest, most accurate,and secure to use<br>**3.** Can be used for Web-Personalizationand Web-Accessibility | **Accuracy:** 98.39%<br>**Precision:** 99.24%<br>**Sensitivity:** 97.83%<br>**F1-Score:** 98.53%<br>**Specificity:** 99.08% |

## 5.2. Cataract Detection Performance

The proposed framework shows the accuracy of 96.90%, precision of 96.92%, and the recall of 97.76% for classifying cataract. The proposed system successfully distinguishes between the healthy and mild versus severe cataract conditions based on the grayscale intensity distribution and opacity analysis in terms of textural features from the lens region.

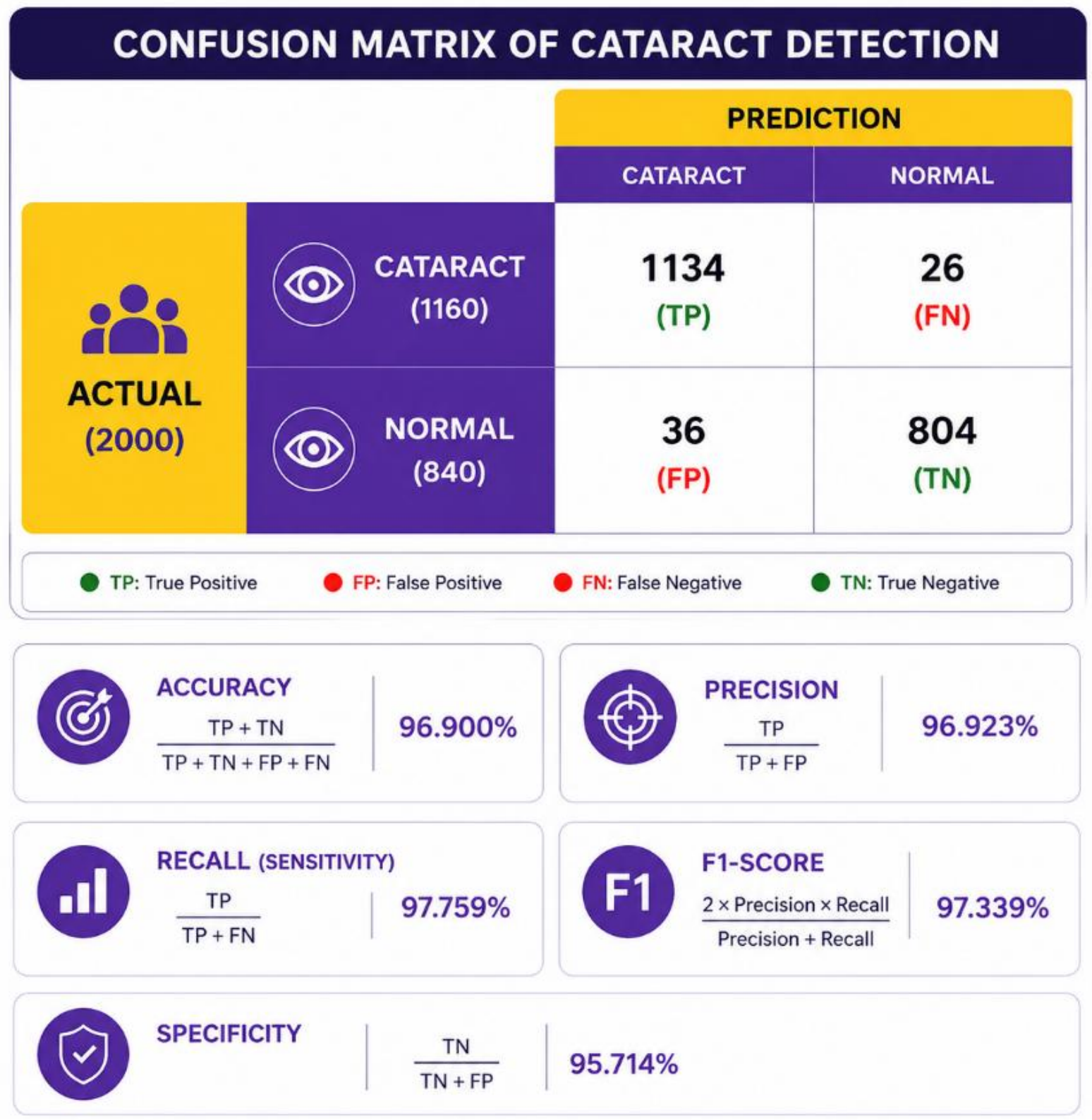

Figure 7: Performance of the model in Cataract detection

The proposed pre-processing and intensity extraction method using the histogram can be used to identify the pattern of protein deposition, which is related to the clouding of the lens, accurately. The proposed intensity thresholding mechanism is able to obtain progressive variations in the lens opacity, which in turn enhances the estimation of cataract severity. The comparative performance of the proposed cataract detection pipeline is shown in Figure 7. Table 2provides a comparative analysis against traditional cataract detection approaches.

Cataract-related visual degradation is a major problem from the accessibility point of view in terms of contrast perception, readability and interaction with modern graphical web interfaces. The suggested detection system can then be used to improve the versatile access systems, which allow for changing the access interface such as contrast, luminance balance, text magnification, glare reduction and edge sharpening. These adaptations can enhance the readability and visual comfort of users with vision impairment due to cataracts.

Table 2: Quantitative Comparison of Cataract Detection Techniques

| Paper | Methodology Used | Application | Accuracy |
|---|---|---|---|
| M. Yang et. al. [25] | Image Preprocessing | Cataract detection from retinal fundus image | no confusion matrix |
| L. Zhang et. al. [26] | Image Preprocessing + DCCN | Cataract detection from retinal fundus image | Accuracy: 89.92% |
| M. S. Junayed et. al. [27] | Image Preprocessing + CatraNet | Cataract detection from retinal fundus image | Accuracy: 99.13% |
| M. Caixinha et. al. [28] | Cataract detection in animal from their ultrasound based retinal fundus image | Accuracy: 99% | |
| I. Khan et. al. [29] | ResNet-50 + Image Processing | Cataract Detection from retinal fundus image | Accuracy: 97.56% |
| J. J. Yang et. al. [30] | Ensemble Learning | Cataract Detection from retinal fundus image | Accuracy: 93.20% |
| H. Wu et. al. [31] | Multi Task Learning | Cataract Detection from retinal fundus image | Accuracy: 96.50% |
| C. Muchibwa et. al. [32] | Survey on different cataract detection from retinal fundus image | | |
| S. M. Khan et. al. [33] | Image Processing +VGG-19 | Cataract detection from retinal fundus image | Accuracy: 97.47% |
| D. Patil et. al. [34] | Survey on different cataract detection technique | | |
| I. Jindal et. al. [35] | Digital Image Processing +Histogram Processing + Area Calculation | Cataract detection from high resolution near eye image | no confusion matrix |
| S. Agustin et. al. [36] | Image Processing + CNN | Cataract detection from high resolution near eye image | Accuracy: 98% |
| H. M. Lopez et. al. [37] | classification of different cataract detection technique based on different parameter | | |
| P. Tripathi et. al. [38] | Image Processing + Pyramid-Net + Iris-pupil segmentation+ | Cataract detection fromhigh resolution Near-Infra-Red eye image | Accuracy: 96.67% |
| T. Ganokratanaa [39] | Image Processing + LeNet Convolutional Neural Network | Cataract detection from high resolution near eye image | Accuracy 96% |
| C. J. Lai et. al. [40] | Image Processing + Convolutional Neural Network | Cataract detection from close monocular eye image | Accuracy: 95% |
| J. Rana et. al. [41] | Android App | Cataract detection from close monocular eye image | No confusion Matrix |
| Proposed Model | Python code based video mining using:<br>1. 478 based facial land mark detection<br>2. Geometrical positioning of pupil w.r.t land mark of respective eye<br>3. Cataract detection based on intensity. | 1. Cataract and its severitydetection from live videos<br>2. cheapest, most accurate,and secure to use<br>3. Can be used forWeb-Personalizationand Web-Accessibility | Accuracy: 96.90%<br>Precision: 96.92%<br>Sensitivity: 97.76%<br>F1-Score:97.34%<br>Specificity: 95.71% |

## 5.3. Accessibility-Aware Adaptive Interaction

The proposed framework extends beyond conventional ocular disease classification by incorporating accessibility-oriented visual impairment inference. Instead of performing diagnosis alone, the system enables personalized accessibility adaptation. Each type of ocular defect has its own set of interaction problems in digital navigation. Cataracts impair contrast sensitivity, clarity and perception of luminance, while squint affects focus stability and continuity of reading. The proposed system could adapt the presentation of the web interface in real-time based on the visual needs of individual users by identifying these impairments. The accessibility adaptation layer can be used to create accessibility adaptations for the following:

- contrast amplification,
- adaptive text scaling,
- optimized component spacing,
- glare reduction,
- simplified layouts,
- guided focus alignment, and
- Visual clutter minimization.

Such accessibility-aware adaptations improve usability, readability, and interaction efficiency for visually impaired users. The proposed framework establishes a connection between AI-based computer vision analysis of the eye and adaptive web tools to improve human-computer interaction in a more inclusive and personalized manner.

### 5.4. Comparative Evaluation

Figure 8presents a comparative analysis between the proposed framework and existing conventional approaches. Experimental results demonstrate that the proposed system outperforms traditional methods in terms of classification accuracy, automation capability, and real-time applicability.

Unlike manual screening procedures and static image-based approaches, the proposed framework performs automated ocular analysis using short-duration video sequences captured through standard laptop or mobile cameras. The ability to process approximately 10-second video recordings significantly improves practical deployment feasibility in remote healthcare and accessibility-oriented environments.

Furthermore, the integration of simultaneous squint and cataract detection within a unified pipeline reduces computational redundancy and enables scalable deployment for intelligent assistive systems. The lightweight video-based implementation also enhances suitability for low-cost and real-time accessibility applications across diverse computing platforms.

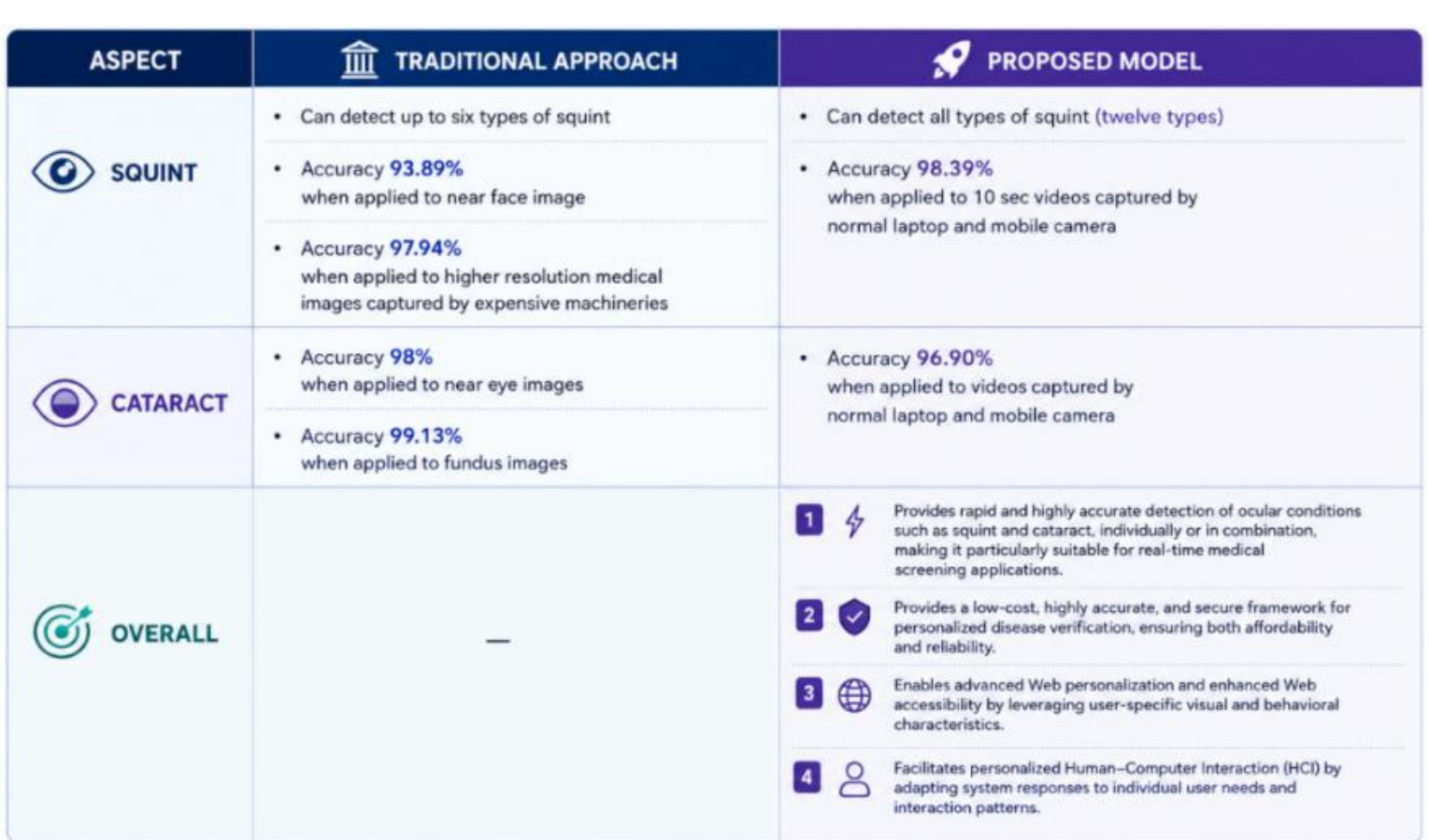

| ASPECT | TRADITIONAL APPROACH | PROPOSED MODEL |
| --- | --- | --- |
| SQUINT | • Can detect up to six types of squint<br>• Accuracy 93.89% when applied to near face image<br>• Accuracy 97.94% when applied to higher resolution medical images captured by expensive machineries | • Can detect all types of squint (twelve types)<br>• Accuracy 98.39% when applied to 10 sec videos captured by normal laptop and mobile camera |
| CATARACT | • Accuracy 98% when applied to near eye images<br>• Accuracy 99.13% when applied to fundus images | • Accuracy 96.90% when applied to videos captured by normal laptop and mobile camera |
| OVERALL | – | 1 Provides rapid and highly accurate detection of ocular conditions such as squint and cataract, individually or in combination, making it particularly suitable for real-time medical screening applications.<br>2 Provides a low-cost, highly accurate, and secure framework for personalized disease verification, ensuring both affordability and reliability.<br>3 Enables advanced Web personalization and enhanced Web accessibility by leveraging user-specific visual and behavioral characteristics.<br>4 Facilitates personalized Human–Computer Interaction (HCI) by adapting system responses to individual user needs and interaction patterns. |

Figure 8: Advantage of proposed Model

## 6. CONCLUSIONS

This study presented a unified video-based framework for automated detection of squint and cataract using computer vision and machine learning techniques. The proposed system integrates geometric ocular alignment analysis with intensity-based lens opacity estimation to perform simultaneous multi-condition ocular screening from image and video data in real time.

Squint detection was performed through spatial analysis of pupil displacement relative to anatomical eye landmarks extracted using a 478-point facial landmark model. The framework successfully classified multiple unilateral and bilateral squint conditions using deterministic geometric rules. Cataract detection was achieved through grayscale intensity distribution and histogram-based opacity analysis of the pupil and lens region, enabling classification of healthy, mild, and severe cataract cases.

Experimental evaluation demonstrated strong diagnostic performance, achieving 98.39\% accuracy for squint detection and 96.90\% accuracy for cataract classification. The lightweight computational architecture allows deployment using standard laptop or mobile cameras, improving accessibility for remote and low-cost ocular screening applications.

Furthermore, the proposed framework supports accessibility-aware visual impairment inference, enabling future integration with adaptive user-interface systems for visually impaired users. Future work will focus on extending the framework toward additional ocular abnormalities and personalized accessibility adaptation mechanisms.

## ACKNOWLEDGEMENTS

We would like to thank the Department of Computer Science and Engineering, Parala Maharaja Engineering College, for providing infrastructure facilities in PARAM SHAVAK High Performance Computing laboratory for conducting our simulation activities. We also extend our heartfelt thanks to Akankshya Kar and Suhani Sukanya Ray for their valuable support in setting up the program.

## FUNDING

No funding was received from any agency, organization, or institution for the conduct of this research.

## CONFLICT OF INTEREST

The authors report no conflicts of interest related to this work.

## AUTHORS

**Amar Ranjan Dash** obtained his B. Tech Comp. Sc. from Biju Patnaik University of Technology, India and M. Tech degree from Berhampur University, India. Currently, He is pursuing his doctoral research. He has Ten research publications to his credit. His research interests include Web Accessibility, Algorithm Optimization, and cloud computing. He is a member of ACM \& IEEE.

**Dr. Manas Ranjan Patra** holds a Ph.D. Degree in Computer Science from the Central University of Hyderabad. He is a Professor in the Department of Computer Science, Berhampur University, India and has been teaching Computer Science for the last 29 years. He was a United Nations Fellow at IIST/UNU, Macao. He has more than 150 research publications to his credit. His research interests include Service Oriented Computing, Applications of Data mining and e-Governance. He has extensively travelled to many countries for presenting research papers, chairing technical sessions and delivering invited talks. He has been a member of Editorial Boards and Programme Committees of many International journals and conferences. He is a member of ACM, CSI and ISTE.

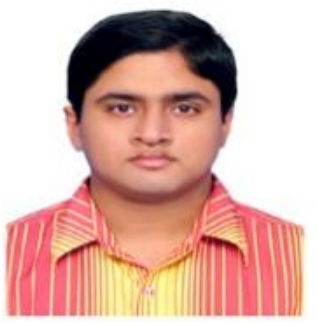

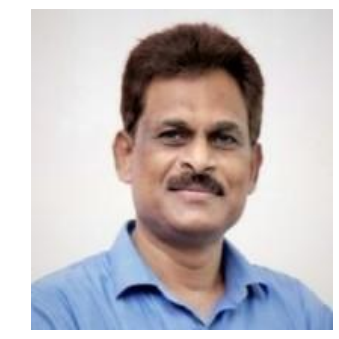